\pdfoutput=1
\documentclass[11pt]{article}

\usepackage{authblk}
\usepackage{acl}
\usepackage{times}
\usepackage{latexsym}
\usepackage{graphicx}
\usepackage[T1]{fontenc}
\usepackage[utf8]{inputenc}
\usepackage{microtype}
\usepackage{smartdiagram}
\usesmartdiagramlibrary{additions}
\usepackage{enumitem}

\title{What is the social benefit of hate speech detection research? A Systematic Review}

\author[1,2]{Sidney G.-J. Wong}
\affil[1]{University of Canterbury, New Zealand}
\affil[2]{Geospatial Research Institute, New Zealand}
\affil[ ]{\texttt{\{sidney.wong\}@pg.canterbury.ac.nz}}

\begin{document}

\maketitle

\begin{abstract}
    While NLP research into hate speech detection has grown exponentially in the last three decades, there has been minimal uptake or engagement from policy makers and non-profit organisations. We argue the absence of ethical frameworks have contributed to this rift between current practice and best practice. By adopting appropriate ethical frameworks, NLP researchers may enable the social impact potential of hate speech research. This position paper is informed by reviewing forty-eight hate speech detection systems associated with thirty-seven publications from different venues.
\end{abstract}

\section{Introduction}
\label{sec:introduction}

    Social impact is a conceptual model used to determine the practice and science of social good factoring: 1) social good domains (including diversity and inclusion; environmental justice and sustainability; and peace and collaboration); 2) unconventional systems of change; and 3) innovative technologies \citep{mor_barak_practice_2020}. Indeed, one area of natural language processing (NLP) which seamlessly unites all three elements of social impact is hate speech detection \citep{hovy_social_2016}. In the last three decades, we have seen an exponential growth into hate speech research with rapid developments in the last decade alone as a result of methodological advancement in NLP \citep{tontodimamma_thirty_2021}. 

    The main contribution of NLP research in combating hate speech is through the development of hate speech detection training data sets. This is because hate speech detection is often treated as a text classification task and the development of hate speech detection systems follow a similar workflow: a) data set collection and preparation; b) feature engineering; c) model training; and lastly d) model evaluation \cite{kowsari_text_2019}. A systematic review of hate speech literature has identified over sixty-nine hate speech detection systems \citep{jahan_systematic_2023}. However, these systems pose a number of ethical challenges and risks to the vulnerable communities they are meant to protect \cite{vidgen_directions_2020}.
    
    As an area of research enquiry, hate speech research is highly productive. For example, the flagship publisher of computational linguistics and natural language processing research, \textit{ACL Anthology}, returned 6,570 results for `hate speech' as of June 2024. This number pales in comparison to the staggering 116,000 publications indexed by Google Scholar. While hate speech research has been purported as a valuable resource in policing anti-social behaviour online \citep{rawat_hate_2024}, some researchers are beginning to question the social benefits of proposed NLP solutions in combating hate speech \citep{parker_is_2023}.
    
    The efforts of NLP researchers are rarely used to combat hate speech. In a review of hate speech policies, the key players in this space were non-profit organisations, social media platforms, and government agencies \citep{parker_is_2023}. Hate speech detection research rarely appear in policy documents. As an example, the most cited hate speech publication had 2,861 citations on Google Scholar \cite{davidson_automated_2017}, but only twice in Overton - a database of policy documents and working papers for 188 countries. The absence of NLP research suggest that methodological innovations are of are incongruent with legal and ethical concerns of this social issue \cite{jin_how_2021}.  

    NLP researchers do not seem to be concerned that their hate speech systems are not being widely applied or implemented. This is because the primary concern in hate speech research is poor model performance which is often attributed noisy training data \citep{arango_hate_2022}. \citet{laaksonen_datafication_2020} critiqued the `datafication' of hate speech research has become an unnecessary distraction for NLP researchers in combating this social issue. This is a well-attested issue in NLP research for positive social impact \cite{diddee_six_2022}
    
    As a relatively new field of academic enquiry \cite{nadkarni_natural_2011}, there remains a paradigmatic rift between current practice and evidence-based best practice. \citet{hovy_social_2016} expressed their concerns on the negative social impacts of NLP research. This is because NLP research was previously immune from research ethics as NLP approaches did not directly involve human subjects. NLP researchers are increasingly aware they are not immune from ethical dilemmas. As an example, recent work have identified racial bias in hate speech systems \cite{davidson_racial_2019}.

    If NLP researchers wish to enable the intended positive social impact of hate speech detection systems, then there must be a re-orientation of how the problem of hate speech detection is conceived from a methods-based problem towards collaborative solution \cite{parker_is_2023}. This view is shared by the broader field of NLP for social good whereby the needs of users and communities are centred over the methods \cite{mukhija_designing_2021}. One proposed approach is to determine the responsibility of NLP solutions and system to consider its broader impact on target users and communities.

\subsection{Responsible Innovation in AI}
\label{sec:responsible_innovation}

    As strands of AI, including NLP, become more intertwined with society, researchers must consciously reflect on the broader ethical implications of their solutions and systems. The \textit{ACM Code of Ethics} exists to support computing professionals \cite{gotterbarn_acm_2018}. However, the perceived opacity in AI research (i.e., poor transparency, explainability, and accountability) led to the recent development of a proposed deliberative framework on responsible innovation \cite{buhmann_towards_2021}. The proposed dimensions of the deliberative framework include:

    \begin{itemize}[nolistsep]
        \item \textit{Responsibility to Prevent Harm}: AI researchers are required to implement risk management strategies in preventing potentially negative outcomes for humans, society, and the environment.
        \item \textit{Obligation to `do good'}: AI researchers and systems are required to improve the conditions for humans, society, and the environment.
        \item \textit{Responsibility to Govern}: AI researchers are stewards of responsible AI systems.
    \end{itemize}

    The conceptual model was influenced by the Principlist approaches in biomedical ethics \cite{beauchamp_principles_2001}. In a similar vein the Principlist principles are used to guide medical professionals in cases of conflict or confusion, the framework was developed to address some of the challenges in AI research at a systemic level. The first dimension corresponds with the Principlist principles of \textit{respect for autonomy} and \textit{non-maleficence}, while the second dimension corresponds with \textit{beneficence} and \textit{justice}.

    When we evaluate existing hate speech research against the proposed deliberative framework, we begin to see where the existing hate speech systems may fall short in terms of social benefits. For example, known biases in hate speech detection systems (e.g., \citealt{davidson_racial_2019}) may further exacerbate inequities of target groups and communities. Additionally, socially or culturally agnostic hate speech systems may offer limited value when applied without considering the sociocultural context of target groups and communities \cite{wong_sociocultural_2024}.

\subsection{Responsible NLP}
\label{sec:rnlp}
  
    Building on the proposed deliberative framework for responsible innovation in AI \cite{buhmann_towards_2021}, \citet{behera_responsible_2023} proposed a conceptual model entitled \textit{Responsible Natural Language Processing} (\textsc{rnlp}) to determine the social benefits of NLP systems throughout its operational life-cycle. The conceptual model was developed from semi-structured interviews with NLP researchers in the health, finance, and retail and e-commerce industries to understand the efficacy of the framework. The NLP researchers found the \textsc{rnlp} a suitable tool for ethical decision making at the structural level.

    \paragraph{Principle 1: Human-Centred Values} NLP systems should respect individual autonomy, diversity, and uphold human rights. NLP systems should not be used to replace cognitive functions (i.e., reasoning, learning, problem solving, perception, and rationality). This also means the perspectives of target communities should be included in the development of the system (i.e., data collection, annotation, deployment). An example of this may involve co-creating NLP informed solutions with target communities \cite{pillai_toward_2023}.

    \paragraph{Principle 2: Transparency} NLP systems should include responsible disclosures especially if a system may have substantial influence on individuals \cite{behera_responsible_2023}. Within a hate speech detection context, disclosures should include a detailed descriptions of the research design including decision-making processes and possible biases or data quality issues. NLP researchers are encouraged to provide data statements profiling participants or annotators and their affiliation to a target group \cite{bender_data_2018}. 

    \paragraph{Principle 3: Well-being} NLP systems should be used to benefit humans, society, and the environment; more importantly, there should be no negative impacts to humans, society, or the environment. These benefits should be explicitly defined and justified. An example of this may involve contextualising the research using the \textit{Researcher Impact Framework} which highlights key achievements in the generation of knowledge, the development of individuals and collaborations, supporting the research community, and supporting broader society \cite{de_moura_rocha_lima_researcher_2022}.

    \paragraph{Principle 4: Privacy and Security} NLP systems should uphold and respect the private rights of individuals. Individuals should not be identified within the system and the system is stored securely. Where appropriate, anonymisation, confidentialisation, or homomorphic encryption should be applied. An example of this may include publishing numerical identifiers of social media posts and not the content without consent \cite{williams_towards_2017}.

    \paragraph{Principle 5: Reliability} NLP systems should operate in a consistent manner (i.e., precise, dependable, and repeatable) in accordance with the intended purpose. An example of this may include publishing code and training data securely as well as relevant model evaluation metrics \cite{resnik_evaluation_2010}. NLP systems should not pose safety risks to individuals.
    
    \paragraph{Principle 6: Fairness} NLP systems should be inclusive and accessible (i.e., user-centric) of marginalised or vulnerable communities. Furthermore, NLP systems should not perpetuate existing prejudice towards marginalised and vulnerable communities. An example of this may include additional assessments for social bias \cite{tan_assessing_2019}. Systems should be deployed on no-code or low-code development platforms as target communities may not have the capability to deploy the system from the source code. Within the context of hate speech detection research, this principle is correlated with \textit{Principle 2: Transparency} and \textit{Principle 8: Accountability}.

    \paragraph{Principle 7: Interrogation} There should be effective and accessible methods that enable individuals to challenge NLP systems. Shared tasks is a useful approach to determine the limitations of the system \cite{parra_escartin_ethical_2017}.

    \paragraph{Principle 8: Accountability} There should be human oversight over the development and deployment of NLP systems throughout various phases of the NLP system life-cycle. Evidence of this principle may include participatory design process with stakeholders \cite{schafer_participatory_2023}; and ethics or internal review board approval obtained.

\subsection{Summary}
\label{sec:summary}

    As target communities continue to experience online hate despite these opaque strategies \citep{burnap_us_2016}, NLP researchers may still play a significant role in unleashing the social impact potential of NLP research - to enable equitable digital inclusion and to close the `digital divide' \citep{norris_digital_2001}. The introduction of the deliberative framework for responsible innovation in AI \cite{buhmann_towards_2021} and the \textit{Responsible NLP} (\textsc{rnlp}) conceptual model \cite{behera_responsible_2023} provide a useful tool to understand the current state of hate speech detection systems. The main contribution of this position paper is a systematic review of existing hate speech detection systems to determine possible areas of improvement with the aim to enable positive social benefits for target groups or communities. We posit the low social impact of hate speech detection research, as evident from the lack of engagement from key stakeholders \cite{parker_is_2023}, may stem from the lack of ethical decision making in the development of these NLP systems.

\section{Analysis}
\label{sec:analysis}

    We retroactively apply the \textsc{rnlp} conceptual model to evaluate the ethical and responsible performance of hate speech systems. Each system is rated on a three-point scale: where there is no evidence (\textit{not met}), some evidence (\textit{partially met}), and good evidence (\textit{met}). While the \textsc{rnlp} evaluates an NLP system in its entirety, we restrict our analysis to the training data sets used to train these systems. As part of our systematic review, we only refer to publicly available publications (or in some instances, pre-prints) and associated data or metadata repository for evidence when evaluating each system.
    
\subsection{Data}
\label{sec:data}

    Even though there are hundreds (possibly thousands) of hate speech detection systems, we have included forty-eight hate speech detection systems which were also reviewed as part of \citet{jahan_systematic_2023}. The list of systems with limited corpus information are presented in the Appendix in Table \ref{tab:2}. For a technical summary of the sample, refer to Tables 11 and 12 in \citet{jahan_systematic_2023}. The systems are associated with thirty-eight publications published between 2016-2020. Furthermore, these hate speech data sets span multiple language conditions.

\section{Results}
\label{sec:results}

    \begin{table}
        \centering
        \begin{tabular}{cccc}
        \hline
        \textbf{\textsc{rnlp}} & \textbf{Met} & \textbf{Partially Met} & \textbf{Not Met} \\
        \hline
        P1 & 4.2\% & 68.8\% & 27.1\% \\
        P2 & 6.3\% & 58.3\% & 35.4\% \\
        P3 & 0.0\% & 33.3\% & 66.7\% \\
        P4 & 39.6\% & 43.8\% & 16.7\% \\
        P5 & 81.3\% & 18.8\% & 0.0\% \\
        P6 & 2.1\% & 33.3\% & 64.6\% \\
        P7 & 52.1\% & 35.4\% & 12.5\% \\
        P8 & 0.0\% & 4.2\% & 95.8\% \\
        \hline
        \end{tabular}
        \caption{Summary table of the systematic review.}
        \label{tab:1}
    \end{table}

    A summary of the results from our systematic evaluation is presented in Table \ref{tab:1}. The evaluation for each hate speech detection system is presented in Table \ref{tab:3} of the Appendix. We do not provide a ranking of the systems in our analysis as the purpose of the systematic review is not to determine the ethical robustness of individual systems. Some systems associated with one publication may appear to have duplicate results as they were developed with a similar methodology.
    
    Most systems (68.8\%) partially met \textit{Principle 1: Well-being} (P1) by explicitly stating the contribution of the system; however, almost a third (27.1\%) of systems did not. Over half (56.3\%) of the systems partially met \textit{Principle 2: Human-Centred Values} (P2) by recruiting manual annotators from relevant sociocultural or linguistic backgrounds; while a third (35.4\%) relied on anonymous crowd-sourcing platforms. Only a third (33.3\%) of systems met \textit{Principle 3: Fairness} (P3) provided a discussion on possible biases, limitations, or data quality issues. The remaining systems did not include a discussion of limitations at all.

    Nineteen systems (39.6\%) met \textit{Principle 4: Privacy and Security} (P4) and twenty-one systems (43.8\%) partially met this principle. The systems which met this principle published de-identified data with a small number stored securely with approval required. Eight systems (16.7\%) did not meet this principle which raises both ethical and legal concerns. Thirty-nine systems (81.3\%) met \textit{Principle 5: Reliability} (P5) while nine systems (18.8\%) partially met this principle. Thirty-one systems (64.6\%) did not meet \textit{Principle 6: Fairness} (P6) as there were no responsible disclosures. The remaining systems (33.3\%) partially met this principle with limited information about the annotators. Over half (52.1\%) of the systems met \textit{Principle 7: Interrogation} (P7). Lastly, the majority (95.8\%) of systems did not meet \textit{Principle 8: Accountability} (P8).

\section{Discussion}
\label{sec:discussion}

    While the systematic review provides useful insights of hate speech detection systems from a structural perspective, it does not provide insights into systemic issues. We therefore organise our discussion using the deliberative framework on responsible innovation in AI \cite{buhmann_towards_2021} to determine the broader ethical implications of the sample of hate speech detection systems as highlighted from our systematic review.

\paragraph{Responsibility to Prevent Harm} 

    The principles associated with this dimension are \textit{Principle 2: Human-Centred Values} and \textit{Principle 6: Transparency}. Based on the systematic review, the sample of systems performed poorly for this dimension. Evidence for \textit{Principle 2: Human-Centred Values} was largely determined by the annotation process of which heavily relied on anonymous crowd-sourcing when labelling the training data sets. Anonymous crowd-sourcing decreases the reliability of the annotated data \cite{ros_measuring_2016}. Manual annotators who may not affiliate with a target group may over generalise linguistic features (i.e., slurs) as hate speech. This dimension requires researchers to implement risk management strategies in preventing negative outcomes for humans, society, and the environment. Only \citet{chung_conan_2019} co-created the detection system alongside target groups and communities. Even though the use of crowd-sourced annotators may seem innocuous from a research design perspective, there is a growing body of evidence that content moderators (in this case manual annotators) are unnecessarily exposed to secondary trauma from harmful content with limited mental health support \cite{spence_content_2024}. This means annotators, whether recruited from within a target group/community or anonymously, may experience harm through the system development process. In terms of evidence for \textit{Principle 6: Transparency}, only one system provided both disclosures and detailed profiles of annotators \cite{alfina_hate_2017}. For example, poor documentation may reinforce existing biases against target communities \cite{arango_hate_2022}.

\paragraph{Obligation to `do good'} 

    The principles associated with this dimension are \textit{Principle 1: Well-being} and \textit{Principle 4: Privacy and Security}. The evidence for \textit{Principle 1: Well-being} was largely determined by the aims and research questions. There was little discussion on the suitability of these systems or the role of target communities or the role of annotators in combating online hate speech. Only two systems, both associated with \citet{chung_conan_2019}, had clear contributions to target communities. While this dimension requires researchers to improve the conditions for humans, society, and the environment, the contributions for most systems were largely methodological and the social benefits were negligible. This reinforces the belief that methodological innovations are incongruent with the social or ethical concerns \cite{jin_how_2021}. In terms of evidence for \textit{Principle 4: Privacy and Security}, this was largely determined by data management practices. The systems which met this principle published de-identified data with a small number stored securely with approval from the researchers required. It is important to note that identifiable social media data contravenes the data use policy of most social media platforms. This means the publication of the availability of these data sets with limited security poses ethical and legal issues. The social benefits of the systems developed resulting from the research should be clear to target groups and communities.

\paragraph{Responsibility to Govern} 

    The remaining four principles are associated with this dimension. The systematic revealed a high degree of polarity in the performance of the principles associated with this dimension. The evidence for \textit{Principle 5: Reliability} was largely determined by the available documentation (i.e., journal article, conference proceeding, or pre-print). We can attribute the high performance of systems in this principle as all associated publications were required to undergo peer-review. The high performance of this principle is in direct contrasts \textit{Principle 6: Reliability} which performed poorly as a majority of systems were not deployed beyond publishing the training data. This meant none of the systems met this principle in its entirety as they are not accessible to target communities. Similarly, all systems performed poorly for \textit{Principle 8: Accountability} as participatory design approaches were non-evident and ethics and internal review board approvals were rarely obtained for these studies. In terms of evidence for \textit{Principle 7: Interrogation}, over half the systems met this principle as the datasets were indexed in Papers with Code or involved with shared tasks which are both effective methods to enable robust interrogation of the systems. Crucially, this is where NLP researchers can enable positive social benefits as this dimension requires researchers to be stewards of responsible AI systems. Social media platforms (such as X (Twitter) and Facebook) remove harmful content using in-house detection algorithms and content moderators \citep{wilson_hate_2021}. This suggests NLP researchers may play a role in challenging these opaque systems and promote transparency, explainability, and accountability of these in-house detection algorithms which continue to fail and expose target groups and communities to hate speech.
    
\section{Conclusion}
\label{sec:conclusion}
    
    While the systematic review cannot determine why there is a lack of engagement from key stakeholders of target groups and communities, the insights on how NLP researchers can improve ethical decision making in the development of hate speech detection systems. Based on the systematic review, NLP researchers working in the field of hate speech detection are consistently meeting the principles of \textit{Principle 5: Reliability}, \textit{Principle 7: Interrogation}, and \textit{Principle 4: Privacy and Security}. The two principles which require the most attention are \textit{Principle 8: Accountability} and \textit{Principle 3: Fairness}. Some of these ethical concerns may be addressed systemically and structurally through the adoption of ethical frameworks (such as \citealt{buhmann_towards_2021} or \citealt{beauchamp_principles_2001}); however, true positive social benefits may only be achieved by working alongside target groups and communities most impacted by this social issue.

\section*{Ethics Statement}
\label{sec:ethics_statement}

    The purpose of this position paper is not to take a punitive view of hate speech detection research, but to determine how NLP researchers can enable ethical research practices in this area. As demographic bias in language models may have unintended downstream impacts on vulnerable and marginalised communities \cite{tan_assessing_2019}; research practices of existing and former hate speech detection systems may also perpetuate unintentional harms on vulnerable and marginalised communities. Even though this position paper is not an NLP system in itself, it does contribute to the development of ethical research practices for NLP systems; therefore, we will use the RNLP \cite{behera_responsible_2023} conceptual model to reinforce current best practice in NLP research.
    
    \paragraph{Principle 1: Well-being} We use the \textit{Researcher Impact Framework} proposed by \citet{de_moura_rocha_lima_researcher_2022} to determine the contributions of this position paper. This position paper contributes to the generation of knowledge in NLP research by evaluating current research practices in hate speech research and the steps needed to enable best practice and ethical research practices. This position supports the development of individuals and the research community by synthesising different ethical conceptual models and frameworks to support best practice in NLP research. While this position paper does not involve vulnerable and marginalised groups, the main contribution of this position paper is to support NLP researchers to effectively address the social issues of broader society by encouraging researcher reflexivity on existing research practices.

    \paragraph{Principle 2: Human-Centred Values} This position paper is a systematic review of existing hate speech detection systems. These are subjective ratings based on the perspectives and experiences of the authors and the ratings have not been automated. We have not used AI assistants in research or writing as this will replace the cognitive functions of the authors. The authors intersect communities often targeted by online hate speech which in turn brings a unique and nuanced perspective on the efficacy of NLP solutions in combating this social issue. The positionality of the authors will be released following anonymous peer-review.
    
    \paragraph{Principle 3: Fairness} This position paper does not perpetuate existing prejudice towards marginalised and vulnerable communities. We are aware that ethical research practice may differ between social, cultural, linguistic, or political affiliations; therefore, we have not associated hate speech systems and their research practices as more or less ethical. We have focused our discussion on social benefits and enabling digital inclusion to avoid taking a deficit approach towards hate speech detection research. We have written this paper in plain language to ensure full accessibility of the content.

    \paragraph{Principle 4: Privacy and Security} This position paper does not contain individually identifying information or examples of hate speech or offensive language. All hate speech detection systems and associated documentation which we have explicitly referenced are available in the public domain.

    \paragraph{Principle 5: Reliability} We have identified no potential risks of this position paper; however, we have not included the complete evaluation of individual systems as this may cause reputational risks for both the developers of the individual systems and the authors of this position paper. As this position paper is largely a qualitative assessment of hate speech detection systems, there are no model evaluation metrics or statistics and we have not included any experimental settings or hyper-parameters.

    \paragraph{Principle 6: Transparency} We have included a brief description of the forty-eight hate speech detection systems which can be located in Table 11 and Table 12 of \citet{jahan_systematic_2023}. We have not involved human subjects or external annotators in our systematic review of hate speech detection systems.

    \paragraph{Principle 7: Interrogation} We encourage other NLP researchers to conduct a similar systematic review based on their own perspectives and experiences. The evaluation with supporting evidence can be made available by contacting the authors.

    \paragraph{Principle 8: Accountability} This position paper does not include human subjects or external annotators; therefore, ethics or internal review board approval have not been sought. However, we encourage NLP researchers working in hate speech detection to contact the authors to discuss the contents of the position paper. We believe there is value in taking a participatory design approach to determine the needs of NLP researchers in hate speech detection to enable ethical research practices.

\section*{Limitations}
\label{sec:limitations}

    This position paper evaluates a sample (48) of existing hate speech detection systems. Naturally, this is not a true reflection of all hate speech detection systems developed or available on the public domain. We suggest elevating this position paper to a bibliometric evaluation of hate speech detection systems to capture the evidence needed to support the claims in this position paper. Furthermore, the qualitative evaluation in this position paper is limited to the perspectives and experiences of the authors; therefore, we do not expect the views expressed in this position paper can be generalised across the NLP research community who may have differing perspectives on best practice ethical research practice which will vary depending on the social, cultural, linguistic, or political affiliations of individuals. This position paper uses one ethical conceptual model and may benefit from the inclusion of other ethical frameworks.

\section*{Acknowledgements}

    The author wants to thank Dr. Benjamin Adams (University of Canterbury | Te Whare Wānanga o Waitaha) and Dr. Jonathan Dunn (University of Illinois Urbana-Champaign) for their supervision and mentorship. The author wants to thank the three anonymous peer reviewers, the area chair, the programme chair, and James Kay for their constructive feedback. Lastly, the author wants to thank Fulbright New Zealand | Te Tūāpapa Mātauranga o Aotearoa me Amerika and their partnership with the Ministry of Business, Innovation, and Employment | Hīkina Whakatutuki for their support through the Fulbright New Zealand Science and Innovation Graduate Award.

\bibliography{references.bib}

\begin{thebibliography}{69}
\providecommand{\natexlab}[1]{#1}

\bibitem[{Alakrot et~al.(2018)Alakrot, Murray, and Nikolov}]{alakrot_dataset_2018}
Azalden Alakrot, Liam Murray, and Nikola~S. Nikolov. 2018.
\newblock \href {https://doi.org/10.1016/j.procs.2018.10.473} {Dataset {Construction} for the {Detection} of {Anti}-{Social} {Behaviour} in {Online} {Communication} in {Arabic}}.
\newblock \emph{Procedia Computer Science}, 142:174--181.

\bibitem[{Albadi et~al.(2018)Albadi, Kurdi, and Mishra}]{albadi_are_2018}
Nuha Albadi, Maram Kurdi, and Shivakant Mishra. 2018.
\newblock \href {https://doi.org/10.1109/ASONAM.2018.8508247} {Are they {Our} {Brothers}? {Analysis} and {Detection} of {Religious} {Hate} {Speech} in the {Arabic} {Twittersphere}}.
\newblock In \emph{2018 {IEEE}/{ACM} {International} {Conference} on {Advances} in {Social} {Networks} {Analysis} and {Mining}}, pages 69--76, Barcelona, Spain. IEEE.

\bibitem[{Alfina et~al.(2017)Alfina, Mulia, Fanany, and Ekanata}]{alfina_hate_2017}
Ika Alfina, Rio Mulia, Mohamad~Ivan Fanany, and Yudo Ekanata. 2017.
\newblock \href {https://doi.org/10.1109/ICACSIS.2017.8355039} {Hate speech detection in the {Indonesian} language: {A} dataset and preliminary study}.
\newblock In \emph{2017 {International} {Conference} on {Advanced} {Computer} {Science} and {Information} {Systems}}, pages 233--238, Bali, Indonesia. IEEE.

\bibitem[{Andrusyak et~al.(2018)Andrusyak, Rimel, and Kern}]{andrusyak_detection_2018}
Bohdan Andrusyak, Mykhailo Rimel, and Roman Kern. 2018.
\newblock \href {http://nlp.fi.muni.cz/raslan/raslan18.pdf#page=85} {Detection of {Abusive} {Speech} for {Mixed} {Sociolects} of {Russian} and {Ukrainian} {Languages}}.
\newblock In \emph{Proceedings in the {Twelfth} {Workshop} on {Recent} {Advances} in {Slavonic} {Natural} {Language} {Processing}}, pages 77--84, Karlova Studánka, Czech Republic. Tribun EU.

\bibitem[{Arango et~al.(2022)Arango, Pérez, and Poblete}]{arango_hate_2022}
Aymé Arango, Jorge Pérez, and Barbara Poblete. 2022.
\newblock \href {https://doi.org/10.1016/j.is.2020.101584} {Hate speech detection is not as easy as you may think: {A} closer look at model validation (extended version)}.
\newblock \emph{Information Systems}, 105:101584.

\bibitem[{Basile et~al.(2019)Basile, Bosco, Fersini, Nozza, Patti, Rangel~Pardo, Rosso, and Sanguinetti}]{basile_semeval-2019_2019}
Valerio Basile, Cristina Bosco, Elisabetta Fersini, Debora Nozza, Viviana Patti, Francisco~Manuel Rangel~Pardo, Paolo Rosso, and Manuela Sanguinetti. 2019.
\newblock \href {https://doi.org/10.18653/v1/S19-2007} {{SemEval}-2019 {Task} 5: {Multilingual} {Detection} of {Hate} {Speech} {Against} {Immigrants} and {Women} in {Twitter}}.
\newblock In \emph{Proceedings of the 13th {International} {Workshop} on {Semantic} {Evaluation}}, pages 54--63, Minneapolis, MN. Association for Computational Linguistics.

\bibitem[{Beauchamp and Childress(2001)}]{beauchamp_principles_2001}
Tom~L. Beauchamp and James~F. Childress. 2001.
\newblock \emph{Principles of {Biomedical} {Ethics}}.
\newblock Oxford University Press.

\bibitem[{Behera et~al.(2023)Behera, Bala, Rana, and Irani}]{behera_responsible_2023}
Rajat~Kumar Behera, Pradip~Kumar Bala, Nripendra~P. Rana, and Zahir Irani. 2023.
\newblock \href {https://doi.org/10.1016/j.techfore.2022.122306} {Responsible natural language processing: {A} principlist framework for social benefits}.
\newblock \emph{Technological Forecasting and Social Change}, 188:122306.

\bibitem[{Bender and Friedman(2018)}]{bender_data_2018}
Emily~M. Bender and Batya Friedman. 2018.
\newblock \href {https://doi.org/10.1162/tacl_a_00041} {Data {Statements} for {Natural} {Language} {Processing}: {Toward} {Mitigating} {System} {Bias} and {Enabling} {Better} {Science}}.
\newblock \emph{Transactions of the Association for Computational Linguistics}, 6:587--604.

\bibitem[{Bohra et~al.(2018)Bohra, Vijay, Singh, Akhtar, and Shrivastava}]{bohra_dataset_2018}
Aditya Bohra, Deepanshu Vijay, Vinay Singh, Syed~Sarfaraz Akhtar, and Manish Shrivastava. 2018.
\newblock \href {https://doi.org/10.18653/v1/W18-1105} {A {Dataset} of {Hindi}-{English} {Code}-{Mixed} {Social} {Media} {Text} for {Hate} {Speech} {Detection}}.
\newblock In \emph{Proceedings of the {Second} {Workshop} on {Computational} {Modeling} of {People}'s {Opinions}, {Personality}, and {Emotions} in {Social} {Media}}, pages 36--41, New Orleans, LA. Association for Computational Linguistics.

\bibitem[{Bretschneider(2016)}]{bretschneider_detecting_2016}
Uwe Bretschneider. 2016.
\newblock \emph{Detecting cyberbullying in online communities}.
\newblock Doctoral, Martin Luther University Halle-Wittenberg, Halle, Germany.

\bibitem[{Buhmann and Fieseler(2021)}]{buhmann_towards_2021}
Alexander Buhmann and Christian Fieseler. 2021.
\newblock \href {https://doi.org/10.1016/j.techsoc.2020.101475} {Towards a deliberative framework for responsible innovation in artificial intelligence}.
\newblock \emph{Technology in Society}, 64:101475.

\bibitem[{Burnap and Williams(2016)}]{burnap_us_2016}
Pete Burnap and Matthew~L. Williams. 2016.
\newblock \href {https://doi.org/10.1140/epjds/s13688-016-0072-6} {Us and them: identifying cyber hate on {Twitter} across multiple protected characteristics}.
\newblock \emph{EPJ Data Science}, 5(1):1--15.

\bibitem[{Chung et~al.(2019)Chung, Kuzmenko, Tekiroglu, and Guerini}]{chung_conan_2019}
Yi-Ling Chung, Elizaveta Kuzmenko, Serra~Sinem Tekiroglu, and Marco Guerini. 2019.
\newblock \href {https://doi.org/10.18653/v1/P19-1271} {{CONAN} - {COunter} {NArratives} through {Nichesourcing}: a {Multilingual} {Dataset} of {Responses} to {Fight} {Online} {Hate} {Speech}}.
\newblock In \emph{Proceedings of the 57th {Annual} {Meeting} of the {Association} for {Computational} {Linguistics}}, pages 2819--2829, Florence, Italy. Association for Computational Linguistics.

\bibitem[{Davidson et~al.(2019)Davidson, Bhattacharya, and Weber}]{davidson_racial_2019}
Thomas Davidson, Debasmita Bhattacharya, and Ingmar Weber. 2019.
\newblock \href {https://doi.org/10.18653/v1/W19-3504} {Racial {Bias} in {Hate} {Speech} and {Abusive} {Language} {Detection} {Datasets}}.
\newblock In \emph{Proceedings of the {Third} {Workshop} on {Abusive} {Language} {Online}}, pages 25--35, Florence, Italy. Association for Computational Linguistics.

\bibitem[{Davidson et~al.(2017)Davidson, Warmsley, Macy, and Weber}]{davidson_automated_2017}
Thomas Davidson, Dana Warmsley, Michael Macy, and Ingmar Weber. 2017.
\newblock \href {http://arxiv.org/abs/1703.04009} {Automated {Hate} {Speech} {Detection} and the {Problem} of {Offensive} {Language}}.
\newblock \emph{arXiv preprint}.
\newblock ArXiv:1703.04009 [cs].

\bibitem[{de~Gibert et~al.(2018)de~Gibert, Perez, García-Pablos, and Cuadros}]{de_gibert_hate_2018}
Ona de~Gibert, Naiara Perez, Aitor García-Pablos, and Montse Cuadros. 2018.
\newblock \href {https://doi.org/10.18653/v1/W18-5102} {Hate {Speech} {Dataset} from a {White} {Supremacy} {Forum}}.
\newblock In \emph{Proceedings of the 2nd {Workshop} on {Abusive} {Language} {Online} ({ALW2})}, pages 11--20, Brussels, Belgium. Association for Computational Linguistics.

\bibitem[{De~Moura Rocha~Lima and Bowman(2022)}]{de_moura_rocha_lima_researcher_2022}
Giovanna De~Moura Rocha~Lima and Sarah Bowman. 2022.
\newblock \href {https://doi.org/10.25546/98474} {Researcher {Impact} {Framework}: {Building} {Audience}-{Focused} {Evidence}-{Based} {Impact} {Narratives}}.
\newblock \emph{Trinity College Dublin}.

\bibitem[{Diddee et~al.(2022)Diddee, Bali, Choudhury, and Mukhija}]{diddee_six_2022}
Harshita Diddee, Kalika Bali, Monojit Choudhury, and Namrata Mukhija. 2022.
\newblock The six conundrums of building and deploying language technologies for social good.
\newblock In \emph{Proceedings of the 5th {ACM} {SIGCAS}/{SIGCHI} {Conference} on {Computing} and {Sustainable} {Societies}}, pages 12--19.

\bibitem[{ElSherief et~al.(2018)ElSherief, Nilizadeh, Nguyen, Vigna, and Belding}]{elsherief_peer_2018}
Mai ElSherief, Shirin Nilizadeh, Dana Nguyen, Giovanni Vigna, and Elizabeth Belding. 2018.
\newblock \href {https://doi.org/10.1609/icwsm.v12i1.15038} {Peer to {Peer} {Hate}: {Hate} {Speech} {Instigators} and {Their} {Targets}}.
\newblock In \emph{Proceedings of the {Twelfth} {International} {AAAI} {Conference} on {Web} and {Social} {Media}}, volume~12, pages 52--61, Palo Alto, CA. Public Knowledge Project.

\bibitem[{Founta et~al.(2018)Founta, Djouvas, Chatzakou, Leontiadis, Blackburn, Stringhini, Vakali, Sirivianos, and Kourtellis}]{founta_large_2018}
Antigoni Founta, Constantinos Djouvas, Despoina Chatzakou, Ilias Leontiadis, Jeremy Blackburn, Gianluca Stringhini, Athena Vakali, Michael Sirivianos, and Nicolas Kourtellis. 2018.
\newblock \href {https://doi.org/10.1609/icwsm.v12i1.14991} {Large {Scale} {Crowdsourcing} and {Characterization} of {Twitter} {Abusive} {Behavior}}.
\newblock In \emph{Proceedings of {theTwelfth} {International} {AAAI} {Conference} on {Web} and {Social} {Media}}, volume~12, Palo Alto, CA. Public Knowledge Project.

\bibitem[{Gao and Huang(2017)}]{gao_detecting_2017}
Lei Gao and Ruihong Huang. 2017.
\newblock \href {https://doi.org/10.26615/978-954-452-049-6_036} {Detecting {Online} {Hate} {Speech} {Using} {Context} {Aware} {Models}}.
\newblock In \emph{Proceedings of the {International} {Conference} {Recent} {Advances} in {Natural} {Language} {Processing}, {RANLP} 2017}, pages 260--266, Varna, Bulgaria. INCOMA Ltd.

\bibitem[{Gomez et~al.(2020)Gomez, Gibert, Gomez, and Karatzas}]{gomez_exploring_2020}
Raul Gomez, Jaume Gibert, Lluis Gomez, and Dimosthenis Karatzas. 2020.
\newblock \href {https://openaccess.thecvf.com/content_WACV_2020/html/Gomez_Exploring_Hate_Speech_Detection_in_Multimodal_Publications_WACV_2020_paper.html} {Exploring {Hate} {Speech} {Detection} in {Multimodal} {Publications}}.
\newblock In \emph{Proceedings of the {IEEE}/{CVF} {Winter} {Conference} on {Applications} of {Computer} {Vision}}, pages 1470--1478.

\bibitem[{Gotterbarn et~al.(2018)Gotterbarn, Brinkman, Flick, Kirkpatrick, Miller, Vazansky, and Wolf}]{gotterbarn_acm_2018}
D.~W. Gotterbarn, Bo~Brinkman, Catherine Flick, Michael~S. Kirkpatrick, Keith Miller, Kate Vazansky, and Marty~J. Wolf. 2018.
\newblock \href {https://www.acm.org/code-of-ethics} {{ACM} code of ethics and professional conduct}.
\newblock Technical report, Association for Computing Machinery.
\newblock Publisher: Association for Computing Machinery.

\bibitem[{Hovy and Spruit(2016)}]{hovy_social_2016}
Dirk Hovy and Shannon~L. Spruit. 2016.
\newblock \href {https://doi.org/10.18653/v1/P16-2096} {The {Social} {Impact} of {Natural} {Language} {Processing}}.
\newblock In \emph{Proceedings of the 54th {Annual} {Meeting} of the {Association} for {Computational} {Linguistics} ({Volume} 2: {Short} {Papers})}, pages 591--598, Berlin, Germany. Association for Computational Linguistics.

\bibitem[{Ibrohim and Budi(2018)}]{ibrohim_dataset_2018}
Muhammad~Okky Ibrohim and Indra Budi. 2018.
\newblock \href {https://doi.org/10.1016/j.procs.2018.08.169} {A {Dataset} and {Preliminaries} {Study} for {Abusive} {Language} {Detection} in {Indonesian} {Social} {Media}}.
\newblock \emph{Procedia Computer Science}, 135:222--229.

\bibitem[{Ibrohim and Budi(2019)}]{ibrohim_multi-label_2019}
Muhammad~Okky Ibrohim and Indra Budi. 2019.
\newblock \href {https://doi.org/10.18653/v1/W19-3506} {Multi-label {Hate} {Speech} and {Abusive} {Language} {Detection} in {Indonesian} {Twitter}}.
\newblock In \emph{Proceedings of the {Third} {Workshop} on {Abusive} {Language} {Online}}, pages 46--57, Florence, Italy. Association for Computational Linguistics.

\bibitem[{Jahan and Oussalah(2023)}]{jahan_systematic_2023}
Md~Saroar Jahan and Mourad Oussalah. 2023.
\newblock \href {https://doi.org/10.1016/j.neucom.2023.126232} {A systematic review of hate speech automatic detection using natural language processing}.
\newblock \emph{Neurocomputing}, 546:126232.

\bibitem[{Jha and Mamidi(2017)}]{jha_when_2017}
Akshita Jha and Radhika Mamidi. 2017.
\newblock \href {https://doi.org/10.18653/v1/W17-2902} {When does a compliment become sexist? {Analysis} and classification of ambivalent sexism using twitter data}.
\newblock In \emph{Proceedings of the {Second} {Workshop} on {NLP} and {Computational} {Social} {Science}}, pages 7--16, Vancouver, Canada. Association for Computational Linguistics.

\bibitem[{Jin et~al.(2021)Jin, Chauhan, Tse, Sachan, and Mihalcea}]{jin_how_2021}
Zhijing Jin, Geeticka Chauhan, Brian Tse, Mrinmaya Sachan, and Rada Mihalcea. 2021.
\newblock \href {https://doi.org/10.18653/v1/2021.findings-acl.273} {How {Good} {Is} {NLP}? {A} {Sober} {Look} at {NLP} {Tasks} through the {Lens} of {Social} {Impact}}.
\newblock In \emph{Findings of the {Association} for {Computational} {Linguistics}: {ACL}-{IJCNLP} 2021}, pages 3099--3113, Online. Association for Computational Linguistics.

\bibitem[{Karim et~al.(2020)Karim, Raja~Chakravarthi, McCrae, and Cochez}]{karim_classification_2020}
Md.~Rezaul Karim, Bharathi Raja~Chakravarthi, John~P. McCrae, and Michael Cochez. 2020.
\newblock \href {https://doi.org/10.1109/DSAA49011.2020.00053} {Classification {Benchmarks} for {Under}-resourced {Bengali} {Language} based on {Multichannel} {Convolutional}-{LSTM} {Network}}.
\newblock In \emph{Proceedings in the 7th {International} {Conference} on {Data} {Science} and {Advanced} {Analytics}}, pages 390--399, Sydney, Australia. IEEE.

\bibitem[{Kowsari et~al.(2019)Kowsari, Jafari~Meimandi, Heidarysafa, Mendu, Barnes, and Brown}]{kowsari_text_2019}
Kamran Kowsari, Kiana Jafari~Meimandi, Mojtaba Heidarysafa, Sanjana Mendu, Laura Barnes, and Donald Brown. 2019.
\newblock \href {https://doi.org/10.3390/info10040150} {Text {Classification} {Algorithms}: {A} {Survey}}.
\newblock \emph{Information}, 10(4):150.

\bibitem[{Kumar et~al.(2018)Kumar, Reganti, Bhatia, and Maheshwari}]{kumar_aggression-annotated_2018}
Ritesh Kumar, Aishwarya~N. Reganti, Akshit Bhatia, and Tushar Maheshwari. 2018.
\newblock \href {https://aclanthology.org/L18-1226} {Aggression-annotated {Corpus} of {Hindi}-{English} {Code}-mixed {Data}}.
\newblock In \emph{Proceedings of the {Eleventh} {International} {Conference} on {Language} {Resources} and {Evaluation} ({LREC} 2018)}, Miyazaki, Japan. European Language Resources Association (ELRA).

\bibitem[{Laaksonen et~al.(2020)Laaksonen, Haapoja, Kinnunen, Nelimarkka, and Pöyhtäri}]{laaksonen_datafication_2020}
Salla-Maaria Laaksonen, Jesse Haapoja, Teemu Kinnunen, Matti Nelimarkka, and Reeta Pöyhtäri. 2020.
\newblock \href {https://doi.org/10.3389/fdata.2020.00003} {The {Datafication} of {Hate}: {Expectations} and {Challenges} in {Automated} {Hate} {Speech} {Monitoring}}.
\newblock \emph{Frontiers in Big Data}, 3.

\bibitem[{Mandl et~al.(2019)Mandl, Modha, Majumder, Patel, Dave, Mandlia, and Patel}]{mandl_overview_2019}
Thomas Mandl, Sandip Modha, Prasenjit Majumder, Daksh Patel, Mohana Dave, Chintak Mandlia, and Aditya Patel. 2019.
\newblock \href {https://doi.org/10.1145/3368567.3368584} {Overview of the {HASOC} track at {FIRE} 2019: {Hate} {Speech} and {Offensive} {Content} {Identification} in {Indo}-{European} {Languages}}.
\newblock In \emph{Proceedings of the 11th {Annual} {Meeting} of the {Forum} for {Information} {Retrieval} {Evaluation}}, pages 14--17, New York, NY. Association for Computing Machinery.

\bibitem[{Mathur et~al.(2018)Mathur, Sawhney, Ayyar, and Shah}]{mathur_did_2018}
Puneet Mathur, Ramit Sawhney, Meghna Ayyar, and Rajiv Shah. 2018.
\newblock \href {https://doi.org/10.18653/v1/W18-5118} {Did you offend me? {Classification} of {Offensive} {Tweets} in {Hinglish} {Language}}.
\newblock In \emph{Proceedings of the 2nd {Workshop} on {Abusive} {Language} {Online} ({ALW2})}, pages 138--148, Brussels, Belgium. Association for Computational Linguistics.

\bibitem[{Mor~Barak(2020)}]{mor_barak_practice_2020}
Michàlle~E. Mor~Barak. 2020.
\newblock \href {https://doi.org/10.1177/1049731517745600} {The {Practice} and {Science} of {Social} {Good}: {Emerging} {Paths} to {Positive} {Social} {Impact}}.
\newblock \emph{Research on Social Work Practice}, 30(2):139--150.

\bibitem[{Mubarak et~al.(2017)Mubarak, Darwish, and Magdy}]{mubarak_abusive_2017}
Hamdy Mubarak, Kareem Darwish, and Walid Magdy. 2017.
\newblock \href {https://doi.org/10.18653/v1/W17-3008} {Abusive {Language} {Detection} on {Arabic} {Social} {Media}}.
\newblock In \emph{Proceedings of the {First} {Workshop} on {Abusive} {Language} {Online}}, pages 52--56, Vancouver, Canada. Association for Computational Linguistics.

\bibitem[{Mukhija et~al.(2021)Mukhija, Choudhury, and Bali}]{mukhija_designing_2021}
Namrata Mukhija, Monojit Choudhury, and Kalika Bali. 2021.
\newblock Designing {Language} {Technologies} for {Social} {Good}: {The} {Road} not {Taken}.
\newblock \emph{arXiv preprint arXiv:2110.07444}.

\bibitem[{Mulki et~al.(2019)Mulki, Haddad, Bechikh~Ali, and Alshabani}]{mulki_l-hsab_2019}
Hala Mulki, Hatem Haddad, Chedi Bechikh~Ali, and Halima Alshabani. 2019.
\newblock \href {https://doi.org/10.18653/v1/W19-3512} {L-{HSAB}: {A} {Levantine} {Twitter} {Dataset} for {Hate} {Speech} and {Abusive} {Language}}.
\newblock In \emph{Proceedings of the {Third} {Workshop} on {Abusive} {Language} {Online}}, pages 111--118, Florence, Italy. Association for Computational Linguistics.

\bibitem[{Nadkarni et~al.(2011)Nadkarni, Ohno-Machado, and Chapman}]{nadkarni_natural_2011}
Prakash~M Nadkarni, Lucila Ohno-Machado, and Wendy~W Chapman. 2011.
\newblock \href {https://doi.org/10.1136/amiajnl-2011-000464} {Natural language processing: an introduction}.
\newblock \emph{Journal of the American Medical Informatics Association}, 18(5):544--551.

\bibitem[{Norris(2001)}]{norris_digital_2001}
Pippa Norris. 2001.
\newblock \href {https://doi.org/10.1017/CBO9781139164887} {\emph{Digital {Divide}: {Civic} {Engagement}, {Information} {Poverty}, and the {Internet} {Worldwide}}}.
\newblock Communication, {Society} and {Politics}. Cambridge University Press, Cambridge.

\bibitem[{Ousidhoum et~al.(2019)Ousidhoum, Lin, Zhang, Song, and Yeung}]{ousidhoum_multilingual_2019}
Nedjma Ousidhoum, Zizheng Lin, Hongming Zhang, Yangqiu Song, and Dit-Yan Yeung. 2019.
\newblock \href {https://doi.org/10.18653/v1/D19-1474} {Multilingual and {Multi}-{Aspect} {Hate} {Speech} {Analysis}}.
\newblock In \emph{Proceedings of the 2019 {Conference} on {Empirical} {Methods} in {Natural} {Language} {Processing} and the 9th {International} {Joint} {Conference} on {Natural} {Language} {Processing} ({EMNLP}-{IJCNLP})}, pages 4675--4684, Hong Kong, China. Association for Computational Linguistics.

\bibitem[{Parker and Ruths(2023)}]{parker_is_2023}
Sara Parker and Derek Ruths. 2023.
\newblock \href {https://doi.org/10.1073/pnas.2209384120} {Is hate speech detection the solution the world wants?}
\newblock \emph{Proceedings of the National Academy of Sciences}, 120(10):e2209384120.

\bibitem[{Parra~Escartín et~al.(2017)Parra~Escartín, Reijers, Lynn, Moorkens, Way, and Liu}]{parra_escartin_ethical_2017}
Carla Parra~Escartín, Wessel Reijers, Teresa Lynn, Joss Moorkens, Andy Way, and Chao-Hong Liu. 2017.
\newblock \href {https://doi.org/10.18653/v1/W17-1608} {Ethical {Considerations} in {NLP} {Shared} {Tasks}}.
\newblock In \emph{Proceedings of the {First} {ACL} {Workshop} on {Ethics} in {Natural} {Language} {Processing}}, pages 66--73, Valencia, Spain. Association for Computational Linguistics.

\bibitem[{Pillai et~al.(2023)Pillai, Griffin, Kronk, and McCall}]{pillai_toward_2023}
Malvika Pillai, Ashley~C. Griffin, Clair~A. Kronk, and Terika McCall. 2023.
\newblock \href {https://doi.org/10.2196/48498} {Toward {Community}-{Based} {Natural} {Language} {Processing} ({CBNLP}): {Cocreating} {With} {Communities}}.
\newblock \emph{Journal of Medical Internet Research}, 25(1):e48498.

\bibitem[{Pitenis et~al.(2020)Pitenis, Zampieri, and Ranasinghe}]{pitenis_offensive_2020}
Zesis Pitenis, Marcos Zampieri, and Tharindu Ranasinghe. 2020.
\newblock \href {https://aclanthology.org/2020.lrec-1.629} {Offensive {Language} {Identification} in {Greek}}.
\newblock In \emph{Proceedings of the {Twelfth} {Language} {Resources} and {Evaluation} {Conference}}, pages 5113--5119, Marseille, France. European Language Resources Association.

\bibitem[{Qian et~al.(2019)Qian, Bethke, Liu, Belding, and Wang}]{qian_benchmark_2019}
Jing Qian, Anna Bethke, Yinyin Liu, Elizabeth Belding, and William~Yang Wang. 2019.
\newblock \href {https://doi.org/10.18653/v1/D19-1482} {A {Benchmark} {Dataset} for {Learning} to {Intervene} in {Online} {Hate} {Speech}}.
\newblock In \emph{Proceedings of the 2019 {Conference} on {Empirical} {Methods} in {Natural} {Language} {Processing} and the 9th {International} {Joint} {Conference} on {Natural} {Language} {Processing} ({EMNLP}-{IJCNLP})}, pages 4755--4764, Hong Kong, China. Association for Computational Linguistics.

\bibitem[{Rawat et~al.(2024)Rawat, Kumar, and Samant}]{rawat_hate_2024}
Anchal Rawat, Santosh Kumar, and Surender~Singh Samant. 2024.
\newblock \href {https://doi.org/10.1002/wics.1648} {Hate speech detection in social media: {Techniques}, recent trends, and future challenges}.
\newblock \emph{WIREs Computational Statistics}, 16(2):e1648.

\bibitem[{Resnik and Lin(2010)}]{resnik_evaluation_2010}
Philip Resnik and Jimmy Lin. 2010.
\newblock \href {https://onlinelibrary.wiley.com/doi/abs/10.1002/9781444324044.ch11} {Evaluation of {NLP} {Systems}}.
\newblock In \emph{The {Handbook} of {Computational} {Linguistics} and {Natural} {Language} {Processing}}, pages 271--295. John Wiley \& Sons, Ltd.

\bibitem[{Rezvan et~al.(2018)Rezvan, Shekarpour, Balasuriya, Thirunarayan, Shalin, and Sheth}]{rezvan_quality_2018}
Mohammadreza Rezvan, Saeedeh Shekarpour, Lakshika Balasuriya, Krishnaprasad Thirunarayan, Valerie~L. Shalin, and Amit Sheth. 2018.
\newblock \href {https://doi.org/10.1145/3201064.3201103} {A {Quality} {Type}-aware {Annotated} {Corpus} and {Lexicon} for {Harassment} {Research}}.
\newblock In \emph{Proceedings of the 10th {ACM} {Conference} on {Web} {Science}}, pages 33--36, New York, NY, USA. Association for Computing Machinery.

\bibitem[{Ribeiro et~al.(2018)Ribeiro, Calais, Santos, Almeida, and Jr}]{ribeiro_characterizing_2018}
Manoel Ribeiro, Pedro Calais, Yuri Santos, Virgílio Almeida, and Wagner~Meira Jr. 2018.
\newblock \href {https://doi.org/10.1609/icwsm.v12i1.15057} {Characterizing and {Detecting} {Hateful} {Users} on {Twitter}}.
\newblock In \emph{Proceedings of the {International} {AAAI} {Conference} on {Web} and {Social} {Media}}, volume~12, Palo Alto, CA. Public Knowledge Project.

\bibitem[{Rizwan et~al.(2020)Rizwan, Shakeel, and Karim}]{rizwan_hate-speech_2020}
Hammad Rizwan, Muhammad~Haroon Shakeel, and Asim Karim. 2020.
\newblock \href {https://doi.org/10.18653/v1/2020.emnlp-main.197} {Hate-{Speech} and {Offensive} {Language} {Detection} in {Roman} {Urdu}}.
\newblock In \emph{Proceedings of the 2020 {Conference} on {Empirical} {Methods} in {Natural} {Language} {Processing}}, pages 2512--2522, Online. Association for Computational Linguistics.

\bibitem[{Roß et~al.(2016)Roß, Rist, Carbonell, Cabrera, Kurowsky, and Wojatzki}]{ros_measuring_2016}
Björn Roß, Michael Rist, Guillermo Carbonell, Benjamin Cabrera, Nils Kurowsky, and Michael Wojatzki. 2016.
\newblock \href {https://doi.org/10.17185/duepublico/42132} {Measuring the {Reliability} of {Hate} {Speech} {Annotations}: the {Case} of the {European} {Refugee} {Crisis}}.
\newblock In \emph{Proceedings in the {Third} {Workshop} on {Natural} {Language} {Processing} for {Computer}-{Mediated} {Communication}}, pages 6--9, Duisburg, Germany. Duisburg-Essen Publications online.

\bibitem[{Sanguinetti et~al.(2018)Sanguinetti, Poletto, Bosco, Patti, and Stranisci}]{sanguinetti_italian_2018}
Manuela Sanguinetti, Fabio Poletto, Cristina Bosco, Viviana Patti, and Marco Stranisci. 2018.
\newblock \href {https://aclanthology.org/L18-1443} {An {Italian} {Twitter} {Corpus} of {Hate} {Speech} against {Immigrants}}.
\newblock In \emph{Proceedings of the {Eleventh} {International} {Conference} on {Language} {Resources} and {Evaluation}}, Miyazaki, Japan. European Language Resources Association (ELRA).

\bibitem[{Schafer et~al.(2023)Schafer, Starbird, and Rosner}]{schafer_participatory_2023}
Joseph~S. Schafer, Kate Starbird, and Daniela~K. Rosner. 2023.
\newblock \href {https://doi.org/10.1145/3563657.3596119} {Participatory {Design} and {Power} in {Misinformation}, {Disinformation}, and {Online} {Hate} {Research}}.
\newblock In \emph{Proceedings of the 2023 {ACM} {Designing} {Interactive} {Systems} {Conference}}, {DIS} '23, pages 1724--1739, New York, NY, USA. Association for Computing Machinery.

\bibitem[{Sigurbergsson and Derczynski(2020)}]{sigurbergsson_offensive_2020}
Gudbjartur~Ingi Sigurbergsson and Leon Derczynski. 2020.
\newblock \href {https://aclanthology.org/2020.lrec-1.430} {Offensive {Language} and {Hate} {Speech} {Detection} for {Danish}}.
\newblock In \emph{Proceedings of the {Twelfth} {Language} {Resources} and {Evaluation} {Conference}}, pages 3498--3508, Marseille, France. European Language Resources Association.

\bibitem[{Spence et~al.(2024)Spence, Bifulco, Bradbury, Martellozzo, and DeMarco}]{spence_content_2024}
Ruth Spence, Antonia Bifulco, Paula Bradbury, Elena Martellozzo, and Jeffrey DeMarco. 2024.
\newblock \href {https://doi.org/10.1089/cyber.2023.0298} {Content {Moderator} {Mental} {Health}, {Secondary} {Trauma}, and {Well}-being: {A} {Cross}-{Sectional} {Study}}.
\newblock \emph{Cyberpsychology, Behavior and Social Networking}, 27(2):149--155.

\bibitem[{Tan and Celis(2019)}]{tan_assessing_2019}
Yi~Chern Tan and L.~Elisa Celis. 2019.
\newblock \href {https://proceedings.neurips.cc/paper/2019/hash/201d546992726352471cfea6b0df0a48-Abstract.html} {Assessing {Social} and {Intersectional} {Biases} in {Contextualized} {Word} {Representations}}.
\newblock In \emph{Advances in {Neural} {Information} {Processing} {Systems}}, volume~32. Curran Associates, Inc.

\bibitem[{Tontodimamma et~al.(2021)Tontodimamma, Nissi, Sarra, and Fontanella}]{tontodimamma_thirty_2021}
Alice Tontodimamma, Eugenia Nissi, Annalina Sarra, and Lara Fontanella. 2021.
\newblock \href {https://doi.org/10.1007/s11192-020-03737-6} {Thirty years of research into hate speech: topics of interest and their evolution}.
\newblock \emph{Scientometrics}, 126(1):157--179.

\bibitem[{Vidgen and Derczynski(2020)}]{vidgen_directions_2020}
Bertie Vidgen and Leon Derczynski. 2020.
\newblock \href {https://doi.org/10.1371/journal.pone.0243300} {Directions in abusive language training data, a systematic review: {Garbage} in, garbage out}.
\newblock \emph{PLOS ONE}, 15(12):e0243300.

\bibitem[{Waseem(2016)}]{waseem_are_2016}
Zeerak Waseem. 2016.
\newblock \href {https://doi.org/10.18653/v1/W16-5618} {Are {You} a {Racist} or {Am} {I} {Seeing} {Things}? {Annotator} {Influence} on {Hate} {Speech} {Detection} on {Twitter}}.
\newblock In \emph{Proceedings of the {First} {Workshop} on {NLP} and {Computational} {Social} {Science}}, pages 138--142, Austin, TX. Association for Computational Linguistics.

\bibitem[{Waseem and Hovy(2016)}]{waseem_hateful_2016}
Zeerak Waseem and Dirk Hovy. 2016.
\newblock \href {https://doi.org/10.18653/v1/N16-2013} {Hateful {Symbols} or {Hateful} {People}? {Predictive} {Features} for {Hate} {Speech} {Detection} on {Twitter}}.
\newblock In \emph{Proceedings of the {NAACL} {Student} {Research} {Workshop}}, pages 88--93, San Diego, CA. Association for Computational Linguistics.

\bibitem[{Wiegand et~al.(2018)Wiegand, Siegel, and Ruppenhofer}]{wiegand_overview_2018}
Michael Wiegand, Melanie Siegel, and Josef Ruppenhofer. 2018.
\newblock \href {https://epub.oeaw.ac.at?arp=0x003a10d2} {Overview of the {GermEval} 2018 {Shared} {Task} on the {Identification} of {Offensive} {Language}}.
\newblock In \emph{Proceedings of {GermEval} 2018, 14th {Conference} on {Natural} {Language} {Processing}}, pages 1--10, Vienna, Austria.

\bibitem[{Williams et~al.(2017)Williams, Burnap, and Sloan}]{williams_towards_2017}
Matthew~L Williams, Pete Burnap, and Luke Sloan. 2017.
\newblock \href {https://doi.org/10.1177/0038038517708140} {Towards an {Ethical} {Framework} for {Publishing} {Twitter} {Data} in {Social} {Research}: {Taking} into {Account} {Users}’ {Views}, {Online} {Context} and {Algorithmic} {Estimation}}.
\newblock \emph{Sociology}, 51(6):1149--1168.

\bibitem[{Wilson and Land(2021)}]{wilson_hate_2021}
Richard Wilson and Molly Land. 2021.
\newblock \href {https://digitalcommons.lib.uconn.edu/law_review/449} {Hate {Speech} on {Social} {Media}: {Content} {Moderation} in {Context}}.
\newblock \emph{Connecticut Law Review}.

\bibitem[{Wong(2024)}]{wong_sociocultural_2024}
Sidney Wong. 2024.
\newblock \href {https://aclanthology.org/2024.c3nlp-1.7} {Sociocultural {Considerations} in {Monitoring} {Anti}-{LGBTQ}+ {Content} on {Social} {Media}}.
\newblock In \emph{Proceedings of the 2nd {Workshop} on {Cross}-{Cultural} {Considerations} in {NLP}}, pages 84--97, Bangkok, Thailand. Association for Computational Linguistics.

\bibitem[{Wulczyn et~al.(2017)Wulczyn, Thain, and Dixon}]{wulczyn_ex_2017}
Ellery Wulczyn, Nithum Thain, and Lucas Dixon. 2017.
\newblock \href {https://doi.org/10.1145/3038912.3052591} {Ex {Machina}: {Personal} {Attacks} {Seen} at {Scale}}.
\newblock In \emph{Proceedings of the 26th {International} {Conference} on {World} {Wide} {Web}}, pages 1391--1399, Republic and Canton of Geneva, CHE. International World Wide Web Conferences Steering Committee.

\bibitem[{Zampieri et~al.(2019)Zampieri, Malmasi, Nakov, Rosenthal, Farra, and Kumar}]{zampieri_predicting_2019}
Marcos Zampieri, Shervin Malmasi, Preslav Nakov, Sara Rosenthal, Noura Farra, and Ritesh Kumar. 2019.
\newblock \href {https://doi.org/10.18653/v1/N19-1144} {Predicting the {Type} and {Target} of {Offensive} {Posts} in {Social} {Media}}.
\newblock In \emph{Proceedings of the 2019 {Conference} of the {North} {American} {Chapter} of the {Association} for {Computational} {Linguistics}: {Human} {Language} {Technologies}, {Volume} 1 ({Long} and {Short} {Papers})}, pages 1415--1420, Minneapolis, MN. Association for Computational Linguistics.

\end{thebibliography}

\section*{Appendix}

    \textit{Continued on the next page}

\begin{table*}
        \footnotesize
        \centering
        \begin{tabular}{lccccc}
        \hline
        \textbf{Citation} & \textbf{Language} & \textbf{Source}& \textbf{Size}& \textbf{Recruitment}& \textbf{Annotators} \\
        \hline
        \citealp{albadi_are_2018} & ar & Twitter & 16,914 & CrowdFlower & - \\
        \citealp{andrusyak_detection_2018} & ru, uk & Youtube & 2,000 & Manual & - \\
        \citealp{bretschneider_detecting_2016} & de & Facebook & 5,836 & Manual & 2 \\
        \citealp{ibrohim_dataset_2018} & id & Twitter & 2,016 & Custom & 20 \\
        \citealp{alakrot_dataset_2018} & ar & Youtube & 15,050 & Mechanical Turk & 3 \\
        \citealp{alfina_hate_2017} & id & Twitter & 713 & Manual & 30 \\
        \citealp{gao_detecting_2017} & en & Fox News & 1,528 & Manual & 2 \\
        \citealp{mubarak_abusive_2017} & ar & Twitter & 1,100 & CrowdFlower & 3 \\
        \citealp{mubarak_abusive_2017} & ar & Al Jazeera & 32,000 & CrowdFlower & 3 \\
        \citealp{jha_when_2017} & en & Twitter & 712 & Manual & 3 \\
        \citealp{jha_when_2017} & en & Twitter & 3,977 & Manual & 3 \\
        \citealp{mulki_l-hsab_2019} & ar & Twitter & 5,846 & Manual & 3 \\
        \citealp{bohra_dataset_2018} & hi-en & Twitter & 4,575 & - & - \\
        \citealp{ibrohim_multi-label_2019} & id & Twitter & 13,169 & Manual & 30 \\
        \citealp{qian_benchmark_2019} & en & GAB & 33,776 & Mechanical Turk & - \\
        \citealp{qian_benchmark_2019} & en & Reddit & 22,324 & Mechanical Turk & - \\
        \citealp{rezvan_quality_2018} & en & Twitter & 24,189 & Manual & 3 \\
        \citealp{ribeiro_characterizing_2018} & en  & Twitter & 4,972 & CrowdFlower & - \\
        \citealp{ros_measuring_2016} & de & Twitter & 469 & Manual & 56 \\
        \citealp{waseem_are_2016} & en & Twitter & 4,033 & CrowdFlower & 2+ \\
        \citealp{waseem_hateful_2016} & en & Twitter & 16,914 & Manual & 4 \\
        \citealp{mathur_did_2018} & hi, en & Twitter & 3,189 & Manual & 3 \\
        \citealp{sanguinetti_italian_2018} & it & Twitter & 1,827 & CrowdFlower & 2+ \\
        \citealp{kumar_aggression-annotated_2018} & hi, en & Facebook & 21,000 & CrowdFlower & 4 \\
        \citealp{kumar_aggression-annotated_2018} & hi, en & Facebook & 18,000 & CrowdFlower & 4 \\
        \citealp{mandl_overview_2019} & en & Twitter, Facebook & 7,005 & Manual & Multiple \\
        \citealp{mandl_overview_2019} & de & Twitter, Facebook & 4,669 & Manual & Multiple \\
        \citealp{mandl_overview_2019} & hi & Twitter, Facebook & 5,983 & Manual & Multiple \\
        \citealp{sigurbergsson_offensive_2020} & da & Multiple & 3,600 & Manual & Multiple \\
        \citealp{wiegand_overview_2018} & de & Twitter & 8,541 & Manual & 3 \\
        \citealp{founta_large_2018} & en & Twitter & 80,000 & CrowdFlower & - \\
        \citealp{karim_classification_2020} & bn & Multiple & 376,226 & Manual & 5 \\
        \citealp{ousidhoum_multilingual_2019} & ar & Twitter & 3,353 & Mechanical Turk & - \\
        \citealp{ousidhoum_multilingual_2019} & en & Twitter & 5,647 & Mechanical Turk & - \\
        \citealp{ousidhoum_multilingual_2019} & fr & Twitter & 4,014 & Mechanical Turk & - \\
        \citealp{pitenis_offensive_2020} & el & Twitter & 4,779 & Manual & 3 \\
        \citealp{rizwan_hate-speech_2020} & ur & Twitter & 10,012 & Manual & 3 \\
        \citealp{zampieri_predicting_2019} & en & Twitter & 14,100 & Figure Eight & - \\
        \citealp{basile_semeval-2019_2019} & es, en & Twitter & 14,100 & Figure Eight & - \\
        \citealp{davidson_automated_2017} & en & Twitter & 24,802 & CrowdFlower & - \\
        \citealp{de_gibert_hate_2018} & en & Stormfront & 9,916 & Manual & 3 \\
        \citealp{elsherief_peer_2018} & en & Twitter & 27,330 & CrowdFlower & - \\
        \citealp{gomez_exploring_2020} & en & Twitter & 149,823 & Mechanical Turk & - \\
        \citealp{wulczyn_ex_2017} & en & Wikipedia & 115,737 & CrowdFlower & - \\
        \citealp{wulczyn_ex_2017} & en & Wikipedia & 100,000 & CrowdFlower & - \\
        \citealp{wulczyn_ex_2017} & en & Wikipedia & 160,000 & CrowdFlower & - \\
        \citealp{chung_conan_2019} & en, fr, it & Facebook & 17,119 & Manual & 20 \\
        \citealp{chung_conan_2019} & en, fr, it & Facebook & 1,288 & Manual & 40 \\
        \hline
        \end{tabular}
        \caption{List of hate speech detection systems surveyed as part of the current systematic evaluation.}
        \label{tab:2}
    \end{table*}

    \begin{table*}
        \footnotesize
        \centering
        \begin{tabular}{lcccccccc}
        \hline
         \textbf{Citation} & \textbf{P1} & \textbf{P2} & \textbf{P3} & \textbf{P4} & \textbf{P5} & \textbf{P6} & \textbf{P7} & \textbf{P8} \\
         \hline
        \citealp{albadi_are_2018} & 1 & 1 & 0 & 2 & 2 & 0 & 0 & 0 \\
        \citealp{andrusyak_detection_2018} & 1 & 0 & 0 & 0 & 2 & 0 & 0 & 0 \\
        \citealp{bretschneider_detecting_2016} & 1 & 2 & 0 & 1 & 2 & 1 & 0 & 0 \\
        \citealp{ibrohim_dataset_2018} & 1 & 1 & 0 & 1 & 2 & 0 & 0 & 0 \\
        \citealp{alakrot_dataset_2018} & 1 & 1 & 1 & 0 & 1 & 1 & 0 & 0 \\
        \citealp{alfina_hate_2017} & 1 & 0 & 1 & 0 & 2 & 2 & 0 & 0 \\
        \citealp{gao_detecting_2017} & 0 & 1 & 0 & 0 & 2 & 0 & 1 & 0 \\
        \citealp{mubarak_abusive_2017} & 0 & 1 & 0 & 1 & 2 & 0 & 1 & 0 \\
        \citealp{mubarak_abusive_2017} & 0 & 1 & 0 & 1 & 2 & 0 & 1 & 0 \\
        \citealp{jha_when_2017} & 0 & 1 & 1 & 2 & 2 & 1 & 1 & 0 \\
        \citealp{jha_when_2017} & 0 & 1 & 1 & 2 & 2 & 1 & 1 & 0 \\
        \citealp{mulki_l-hsab_2019} & 0 & 1 & 1 & 0 & 2 & 1 & 1 & 0 \\
        \citealp{bohra_dataset_2018} & 1 & 0 & 0 & 2 & 2 & 0 & 1 & 0 \\
        \citealp{ibrohim_multi-label_2019} & 1 & 1 & 0 & 1 & 2 & 0 & 1 & 0 \\
        \citealp{qian_benchmark_2019} & 1 & 0 & 0 & 1 & 2 & 0 & 1 & 0 \\
        \citealp{qian_benchmark_2019} & 1 & 0 & 0 & 1 & 2 & 0 & 1 & 0 \\
        \citealp{rezvan_quality_2018} & 1 & 1 & 0 & 2 & 1 & 0 & 1 & 0 \\
        \citealp{ribeiro_characterizing_2018} & 1 & 0 & 0 & 2 & 2 & 0 & 1 & 0 \\
        \citealp{ros_measuring_2016} & 1 & 1 & 0 & 1 & 2 & 0 & 1 & 0 \\
        \citealp{waseem_are_2016} & 1 & 1 & 0 & 2 & 2 & 0 & 1 & 0 \\
        \citealp{waseem_hateful_2016} & 1 & 1 & 0 & 2 & 2 & 0 & 1 & 0 \\
        \citealp{mathur_did_2018} & 1 & 1 & 1 & 2 & 2 & 1 & 1 & 0 \\
        \citealp{sanguinetti_italian_2018} & 1 & 1 & 1 & 2 & 2 & 1 & 1 & 0 \\
        \citealp{kumar_aggression-annotated_2018} & 0 & 1 & 0 & 2 & 1 & 0 & 2 & 0 \\
        \citealp{kumar_aggression-annotated_2018} & 0 & 1 & 0 & 2 & 1 & 0 & 2 & 0 \\
        \citealp{mandl_overview_2019} & 0 & 1 & 0 & 0 & 1 & 0 & 2 & 0 \\
        \citealp{mandl_overview_2019} & 0 & 1 & 0 & 1 & 1 & 0 & 2 & 0 \\
        \citealp{mandl_overview_2019} & 0 & 1 & 0 & 1 & 1 & 0 & 2 & 0 \\
        \citealp{sigurbergsson_offensive_2020} & 0 & 1 & 0 & 1 & 2 & 0 & 2 & 0 \\
        \citealp{wiegand_overview_2018} & 0 & 1 & 1 & 0 & 1 & 1 & 2 & 0 \\
        \citealp{founta_large_2018} & 1 & 0 & 0 & 2 & 2 & 0 & 2 & 0 \\
        \citealp{karim_classification_2020} & 1 & 1 & 0 & 1 & 2 & 0 & 2 & 0 \\
        \citealp{ousidhoum_multilingual_2019} & 1 & 0 & 0 & 1 & 2 & 0 & 2 & 0 \\
        \citealp{ousidhoum_multilingual_2019} & 1 & 0 & 0 & 1 & 2 & 0 & 2 & 0 \\
        \citealp{ousidhoum_multilingual_2019} & 1 & 0 & 0 & 1 & 2 & 0 & 2 & 0 \\
        \citealp{pitenis_offensive_2020} & 1 & 1 & 0 & 1 & 2 & 0 & 2 & 0 \\
        \citealp{rizwan_hate-speech_2020} & 1 & 1 & 0 & 1 & 2 & 0 & 2 & 0 \\
        \citealp{zampieri_predicting_2019} & 1 & 0 & 0 & 2 & 1 & 0 & 2 & 0 \\
        \citealp{basile_semeval-2019_2019} & 1 & 0 & 1 & 2 & 2 & 1 & 2 & 0 \\
        \citealp{davidson_automated_2017} & 1 & 1 & 1 & 0 & 2 & 1 & 2 & 0 \\
        \citealp{de_gibert_hate_2018} & 1 & 1 & 1 & 1 & 2 & 1 & 2 & 0 \\
        \citealp{elsherief_peer_2018} & 1 & 0 & 1 & 2 & 2 & 1 & 2 & 0 \\
        \citealp{gomez_exploring_2020} & 1 & 0 & 1 & 1 & 2 & 1 & 2 & 0 \\
        \citealp{wulczyn_ex_2017} & 1 & 0 & 1 & 2 & 2 & 1 & 2 & 0 \\
        \citealp{wulczyn_ex_2017} & 1 & 0 & 1 & 2 & 2 & 1 & 2 & 0 \\
        \citealp{wulczyn_ex_2017} & 1 & 0 & 1 & 2 & 2 & 1 & 2 & 0 \\
        \citealp{chung_conan_2019} & 2 & 2 & 0 & 1 & 2 & 0 & 2 & 1 \\
        \citealp{chung_conan_2019} & 2 & 2 & 0 & 1 & 2 & 0 & 2 & 1 \\

        \hline
        \end{tabular}
        \caption{The systematic evaluation of hate speech detection systems. We have indicated for each system where there is no evidence (0), some evidence (1), and good evidence (2) for each principle.}
        \label{tab:3}
    \end{table*}

\end{document}